\documentclass{Interspeech2024}

\usepackage{subfigure}

\interspeechcameraready

\title{Multimodal Segmentation for Vocal Tract Modeling}

\name[]{Rishi}{Jain$^*$}
\name[]{Bohan}{Yu$^*$}
\name[]{Peter}{Wu}
\name[]{Tejas}{Prabhune}
\name[]{Gopala}{Anumanchipalli}

\address{
  University of California, Berkeley, United States} %\\
\email{rishiraij@berkeley.edu, ybhtim@berkeley.edu} %, third@companyB.ai}

 % TODO

\begin{document}

\maketitle

\def\thefootnote{*}\footnotetext{These authors contributed equally to this work}\def\thefootnote{\arabic{footnote}}

\begin{abstract}
%igh quality speech-to-avatar synthesis is important for reliable facial rendering in graphics, and also has applications in second language learning, phonetics, and VR for embodying paralyzed patients. We propose a high-resolution avatar animation system using Real-Time Magnetic Resonance Imaging (RT-MRI) as an intermediary for tracking active articulators during naturalistic speaking. Through multimodal RT-MRI segmentation, we demonstrate effective labeling of data from 75 unlabeled speakers, inversion from speech-to-MRI trajectories, and direct real-time mapping to a custom 3D model [of tongue, lips jaw etc] in Unreal Engine. {\textcolor{red}{Note: Right now segmentation is unet which is not multi-modal.}}[Our approach generalizes to avatar animation of unseen speakers, using only speech to drive the 3D facial avatar with a latency of xxx milliseconds and accuracy of xxx.]
Accurate modeling of the vocal tract is necessary to construct articulatory representations for interpretable speech processing and linguistics. However, vocal tract modeling is challenging because many internal articulators are occluded from external motion capture technologies. Real-time magnetic resonance imaging (RT-MRI) allows measuring precise movements of internal articulators during speech, but annotated datasets of MRI are limited in size due to time-consuming and computationally expensive labeling methods. We first present a deep labeling strategy for the RT-MRI video using a vision-only segmentation approach. We then introduce a multimodal algorithm using audio to improve segmentation of vocal articulators. Together, we set a new benchmark for vocal tract modeling in MRI video segmentation and use this to release labels for a 75-speaker RT-MRI dataset, increasing the amount of labeled public RT-MRI data of the vocal tract by over a factor of 9. The code and dataset labels can be found at \url{rishiraij.github.io/multimodal-mri-avatar/}.

\end{abstract}
\section{Introduction}
\label{sec:intro}

Vocal tract modeling is an essential technology in many applications including facial animation, naturalistic speaking avatars, speaker modeling, and second language 
pronunciation learning \cite{sla2015, emavisual2010, ultra2008, bravo2023, languagetrain, languagetrain2}.
In fact, popular self-supervised speech representations inherently learn features correlated with articulators \cite{probing}. 
Modeling is also necessary in healthcare applications such as 
brain-computer interfaces for communication \cite{bravo2023, gopalanature2019} and treating speech disfluencies \cite{stutter, Richard_2021_WACV}. Methods of external motion capture cannot 
record precise and accurate vocal tract movements for occluded
articulators. Thus, the inner mouth is often poorly represented
or neglected in multimedia approaches to motion capture-based 
facial animation \cite{salvador2022}. Popular approaches to solving the issue of inner mouth occlusion 
include electromagnetic articulography (EMA) and electromyography (EMG)
as models for the vocal tract. However, these methods only contain
a small subset of articulatory features \cite{ema_old, gaddy-klein-2021-improved}.

A more comprehensive approach uses Real-Time Magnetic Resonance Imaging (RT-MRI) of the vocal tract \cite{rtmri_articulation_example}. 
This technology offers audio-aligned videos of internal
and external articulators that are not measurable by other articulatory
representations. When tested on downstream speech-related tasks, 
RT-MRI has been shown to more reliably and completely model the vocal 
tract in comparison to EMA \cite{wu2023mrisynth}. For example, MRI representations distinguish
between oral vowels (lowered velum) and nasal vowels (raised 
velum), while EMA does not track the velum at all.
However, current state-of-the-art
labeling methods for extracting interpretable features from these videos
are time-consuming, computationally expensive, and prone to errors
\cite{iter_seg}. Therefore, only a small
amount of vocal tract RT-MRI data is labeled \cite{rtmri2014}. As a result, current work using real-time articulatory MRI falls into two broad categories: (1) methods which rely on the previously extracted articulator segmentations \cite{wu2023mrisynth, stutter}, or (2) models which directly work with RT-MRI videos but do not contain an interpretable intermediate representation \cite{yu2021, otani23mri}. To address the scarcity of publicly-available articulatory segmentations for RT-MRI, we propose:
\begin{itemize}
    \item{A vision-based fully-convolutional neural network \cite{u_net} for speaker-independent vocal tract boundary segmentation.}
    \item{A multimodal Transformer model which additionally includes the speech waveform to set a new benchmark for vocal tract RT-MRI segmentation.}
    \item{Labels for the 75-speaker Speech MRI Open Dataset \cite{usc75} containing 
        over 20 hours of vocal tract RT-MRI data for 75 speakers diverse in
        age, gender, and accent.}
\end{itemize}

\section{Datasets}
\label{sec:datasets}

\subsection{USC-TIMIT Dataset}

We use the labeled 8-speaker RT-MRI 
USC-TIMIT dataset of the vocal tract described in \cite{rtmri2014} for training. 
Subjects were instructed to read phonetically-diverse sentences out loud at a natural speaking
rate while laying supine in an MRI scanner. A four-channel upper airway
receiver coil array was used for signal reception, which was
processed to reproduce $84 \times 84$ pixel midsaggital MRI videos capturing 
lingual, labial, and jaw motion, and velum, pharynx, and larynx articulations.
These videos are collected at 83.33 Hz. We start with the 170 representative points from \cite{rtmri2014} to represent vocal tract air-tissue boundary segmentations. Of these 170 points, we 
take the subset of 95 points (190 $x$ and $y$ coordinates) that has been 
determined to be most vital for speech tasks
in Wu \textit{et al.} \cite{wu2023mrisynth}. All RT-MRI video in the USC-TIMIT dataset is accompanied by existing 
articulator points extracted using the baseline algorithm described further in Section~\ref{ssec:baseline}. We use these point labels as training targets for the other segmentation methods described in Section~\ref{sec:models}. Paired with these trajectories is the 16kHz
speech data (resampled from original 20kHz) 
corresponding to the spoken audio during the RT-MRI scan.
Following previous articulatory MRI work, we further enhance this audio using Adobe Podcast to reduce 
reverbation \cite{wu2023mrisynth}. For training, we use 7 of the 8 speakers (roughly 66 minutes of RT-MRI video) and leave out the remaining speaker as ``unseen''.

\subsection{Speech MRI Open Dataset}
\label{sec:newdataset}

The Speech MRI Open Dataset \cite{usc75} is a diverse 75-speaker dataset 
that provides 20 hours of raw multi-coil RT-MRI videos of the vocal 
tract during articulation, aligned with corresponding speech. Such a large, rich dataset 
can help solve many open problems in fields 
related to phonetics, spoken language, and vocal
articulation. However, unlike the USC-TIMIT dataset, the data does not include labeled MRI 
feature points tracked over time.
\section{Models and Training}
\label{sec:models}

\subsection{Frequency-domain Gradient Descent Baseline}
\label{ssec:baseline}

The existing algorithm for articulatory RT-MRI segmentation \cite{rtmri2014} relies on hand-traced 
air-tissue boundaries for the first frame of every video. This is followed by nonlinear 
optimization in the frequency space of subsequent frames, requiring 20 minutes to converge for a 
single frame using gradient descent. This procedure is also prone to mislabeling and requires 
human supervision, making it expensive to run. Because each frame is optimized independently, it often results in jitter, or high-frequency perturbations, for individual articulator points across consecutive frames. As this is the only existing algorithm for articulatory RT-MRI labeling, the outputs of this model are used as the ``ground truth'' training targets for the following models, and the algorithm will be referred to as the ``baseline'' algorithm.

\subsection{Heatmap U-Net}
\label{ssec:unet}

\begin{figure}
\includegraphics[width=\linewidth]{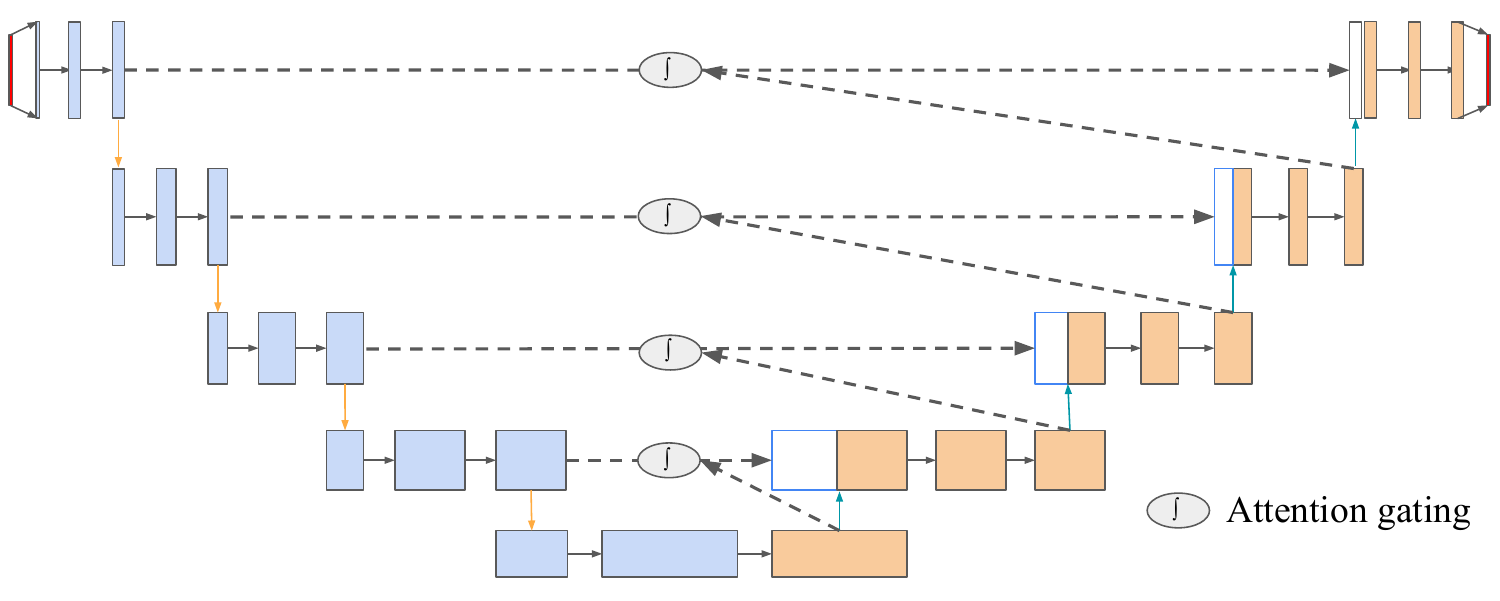}
\caption{The attention U-Net model. Dotted lines represent the paths of attention gating in contracting/expanding layers.}
\label{fig:unet}
\end{figure}

The U-Net \cite{u_net}, a residual fully-convolutional neural network, has historically performed well on low resolution medical images, especially when training data is limited. Because labeled data was only originally available from eight speakers, this architecture was a natural fit. Input MRI frames were padded to a spatial dimension of 96 by 96 and subsequently reduced in the spatial dimension by a factor of two in each layer of the contracting path before expanding. Of the spatial features, the key articulators only occupy a subset of the space. For this reason, we apply attention gating following the Attention U-Net \cite{attention_unet} with the modification of using additive attention as opposed to multiplicative, visualized in Figure~\ref{fig:unet}. While minimally increasing complexity, the model learns to suppress the components of the signal which are not important for the labeling task.
% learned a spatial weighting map on the residual connection to effectively suppress  using an attention gating mechanism, and introduced normalization layers similar to the Attention U-Net . Adding attention gating minimally increased model complexity. The architecture is visualized in .

We trained this model on approximately 90 minutes of labeled midsaggital RT-MRI video from 7 speakers for a total of 6 epochs. The model outputs a 96 by 96 grid for each of the 95 articulatory points. Each of the target keypoints were modeled as 2-dimensional isotropic Gaussian distributions over the 96 by 96 spatial grid with a standard deviation of 2 pixels. For generating keypoint locations from the output heatmaps, we took a weighted average of the $k$ pixels with the highest output values, where the best $k$ was found experimentally to be 25. During training, we also applied random affine transformations to frames and the corresponding annotations to promote generalization to unseen speakers. 

Typically, the pixelwise mean squared error loss, also known as L2 loss, is used for heatmap regression tasks, but we also introduce using the Kullback–Leibler (KL) divergence between the output and target grids in which each output grid is restricted to a 2-dimensional probability distribution using a softmax nonlinearity. To our knowledge, this training objective has not been used for heatmap regression in medical imaging in the past, but guides the model into producing an output that also appears Gaussian in nature and is intuitively well suited for measuring the difference in the two probability distributions. 

In addition, articulator points have varying degrees of movement (standard deviation) and importance in speech production. In this context, articulator ``importance'' is determined by the effect that dropping the articulator has on downstream speech synthesis. Both the importance and standard deviation were calculated using the 7 training speakers by previous works \cite{wu2023mrisynth}. We multiply the standard deviation and importance of each point to determine its weighting in the combined loss. This articulatory weighting emphasizes the importance of points that show significant movement and are important to speech production over those which show minimal movement or have been found to be less essential.

% Of the spatial features, the key articulators only occupy a subset of the space. For this reason, we learn a spatial weighting map on the residual connection to effectively suppress the components of the signal which are not important for speech. To do this, we introduce an attention gating mechanism, similar to the Attention U-Net \cite{attention_unet}. This adds minimal additional complexity, but filters the target spatial area from the residual activations of the corresponding contraction block before appending to the expansion block. The spatial weighting map is produced by information from both the residual activations of the contraction block (pixel vector $x$) and the previous expansion block (gating vector $g$), allowing it to take into account low-level spatial information and downstream high-level semantic context. Specifically, we pass $x$ and $g$ through 1x1 convolutional layers to match the spatial dimensions and use additive attention with an additional convolution to reduce to one channel. This is our learned spatial weighting which is passed through a sigmoid activation and upsampled back to the spatial dimension of our pixel vector $x$ and multiplied to produce the final weighted activations that we then concatenate as part of the residual connection.

\subsection{Multimodal Transformer}

\begin{figure}
\includegraphics[width=\linewidth]{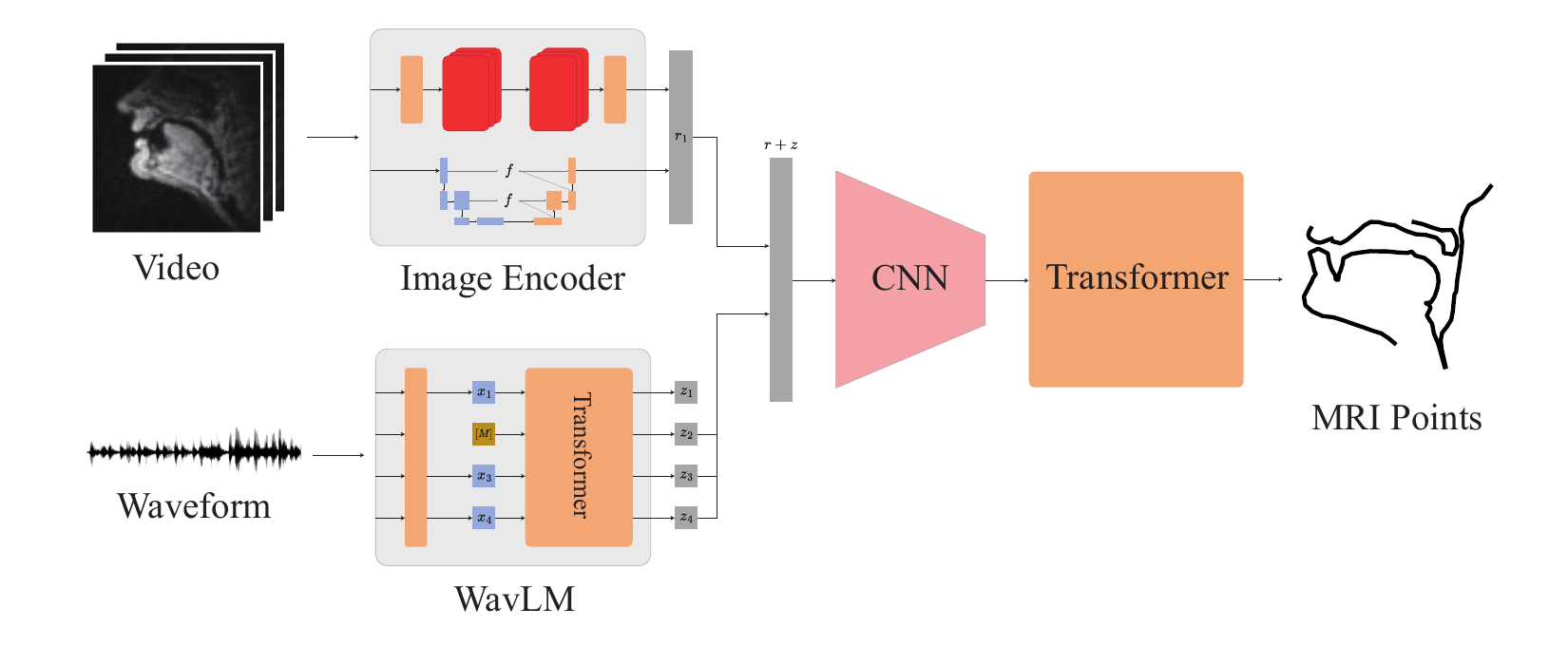}
\caption{Architecture of the multimodal segmentation model.}
\label{fig:multimodal}
\end{figure}

Using the U-Net model as a pretrained convolutional input, we 
further explored joint point tracking methods. To ensure tracks remain smooth, we applied a temporal Gaussian low-pass filter independently for each point of the U-Net output.
We also tried using a convolutional 
LSTM as in \cite{mriseg2020} (CLSTM) and 
a Transformer. The CLSTM, previously used in MRI video 
segmentation \cite{yu2021}, applies a 2-layer LSTM to the 
predicted U-Net outputs, trained on 
speech from the same 7 USC-TIMIT speakers. The Transformer 
similarly used the U-Net points from each timestep, with an 
additional positional encoding. Additionally, we experimented with adding optical flow, 
Kalman filtering, and Lucas Kanade to improve temporal point tracking \cite{lucas, kalman}. Both the CRNN and the Transformer methods did not achieve equal 
or better performance than smoothed U-Net tracks on MRI videos 
of unseen speakers, reinforcing the fact that articulatory MRI tracking 
is fundamentally different than other traditional video-only 
tracking problems.

We subsequently experimented with multimodal models
for feature extraction, using representations from
video frames and speech waveforms. For video frames, we used the output of the frozen U-Net model described in Section~\ref{ssec:unet} and also experimented with other image representation models including ResNet \cite{he2015deep} and ConvNeXt \cite{liu2022convnet}. To represent audio, we used the 10th layer of WavLM \cite{chen2022wavlm} to derive speech representations. The two representations were then concatenated as input to a Transformer prepended with three residual convolutional blocks as seen in Figure~\ref{fig:multimodal}. 
Additionally, we experimented with an audio-only segmentation model (articulatory inversion) using the same WavLM and Transformer methods. The Transformer models were trained on the
speech data from the same 7 of 8 USC-TIMIT speakers as in Section~\ref{ssec:unet}. Using multi-task
learning, the Transformer experiments output MRI trajectories and pitch simultaneously, optimized using weighted L1 loss. 

\section{Results}
\label{sec:results}

We performed quantitative evaluations of both our vision-based and 
multimodal vocal tract segmentation approaches. The segmentations were then used to add articulatory 
labels to RT-MRI from 75 previously-unlabeled speakers.
Using this data as a multimodal pretraining approach, the different segmentations 
were further used for a downstream speech task to measure how well speech features were captured 
by different segmentation methods. Finally, we conducted a qualitative hypothesis test using our best method. 

\subsection{Vision-only U-Net}
\label{ssec:unet_results}

The first experiment compared L2 (mean squared error) loss against our new pixel-wise KL-divergence loss with and without articulatory weighting for the U-Net model. This was evaluated using the root mean squared error (RMSE) of the predicted x-y points for the 95 articulator points on an unseen speaker. The results in Table~\ref{tab:unet_loss} demonstrate that the KL-divergence loss is better suited for low-resolution point recognition for air-tissue boundary segmentation. As RMSE and MSE have the same convergence point, articulatory weighting predictably appears worse using this metric. However, manual inspection reveals that most of this error can be attributed to shifts in less phonologically important articulators such as the hard palate, with significant improvement on the more important articulators.

% \begin{table}
%     \centering
%     \caption{Comparison of the root mean squared error of the models trained using L2 loss, KL-divergence loss, and KL-divergence loss with articulatory weighting. More details are available in Section~\ref{ssec:unet_results}.}
%     \begin{tabular}{ccc}
%     \toprule
%    %\hline
%       Loss       & RMSE\\  \hline
%   MSE (L2)       &7.33\\
%   KL-div       &3.74\\
%   KL-div + Weighting       &3.92\\
%     %\hline
%     \bottomrule
%     \end{tabular}
%     \label{tab:unet_loss}
% \end{table}

\begin{table}
    \centering
    \caption{Comparison of the root mean squared error of the U-Net models trained using L2 loss, KL-divergence loss, and KL-divergence loss with articulatory weighting. More details are available in Section~\ref{ssec:unet_results}.}
    \begin{tabular}{ccc}
    \toprule
    
    \textbf{Loss} & \textbf{RMSE} \\ \midrule
    
  MSE (L2)       &7.33\\ 
  KL-div       &3.74\\ 
  KL-div + Weighting       &3.92\\ 
    %\hline
    \bottomrule
    \end{tabular}
    \label{tab:unet_loss}
\end{table}

\subsection{Labeling the Speech MRI Open Dataset}

The vision-based U-Net above was used to provide labels to RT-MRI video for the 75 speakers in the Speech MRI Open Dataset \cite{usc75}. Outputs from this model were subsequently run through a temporal Gaussian low-pass filter, which was applied independently for each articulator x-y point and used to provide video and audio aligned MRI trajectories.

\begin{figure}[!htb]
    \centering
    \includegraphics[width=0.8\linewidth]{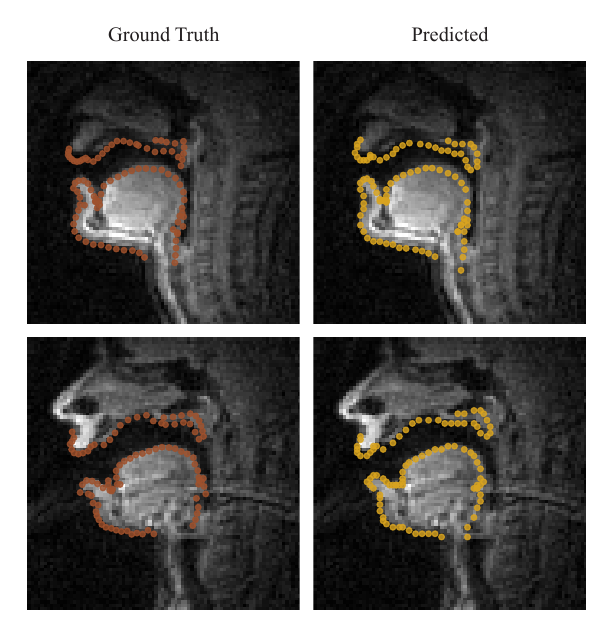}
    \caption{Two representative examples of predicted MRI points (\textit{right}) compared to 
    expert hand labels (\textit{left}). The examples are spoken by unseen Female (\textit{bottom}) and Male (\textit{top}) speakers in the Speech MRI Open Dataset.}
    \label{fig:Labeled75}
\end{figure}

In Figure \ref{fig:Labeled75}, we highlight the generalization
of the U-Net model on unseen speakers,
allowing us to expand the amount of labeled RT-MRI video to over 20 hours
across 83 total speakers. Qualitatively, the predicted segmentations 
closely follow the MRI segments,
achieving high quality labeling for unseen speakers. As part of 
this paper, we also present this labeling for use in future 
downstream speech tasks, increasing the amount of 
publically-available labeled articulatory RT-MRI data by over a factor of \textbf{9}. The labels are available at \url{rishiraij.github.io/multimodal-mri-avatar/}.
% labels are available at [\textit{to ensure author anonymity, the link to the resource will be added after the review process}].

\subsection{Comparison with Multimodal Transformer}

\begin{figure*}[!htb]
    \centering
    \subfigure{\includegraphics[width=0.4\textwidth]{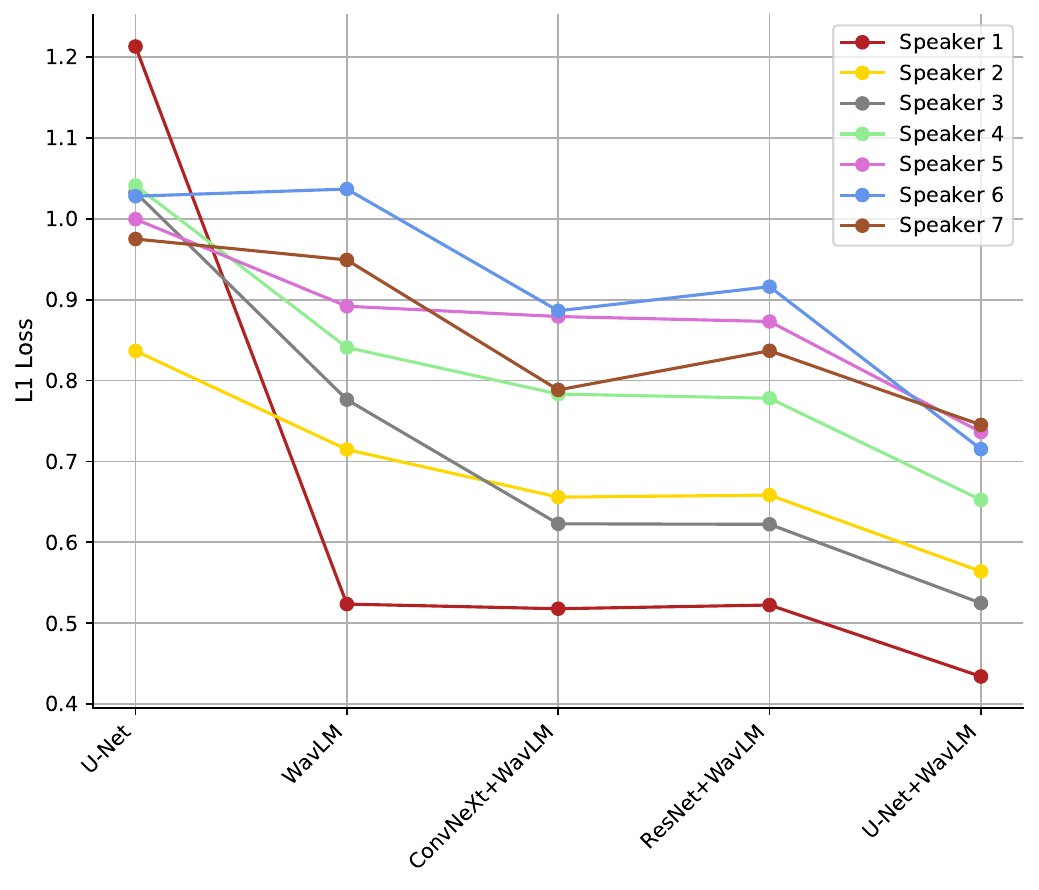}}
    \subfigure{\includegraphics[width=0.4\textwidth]{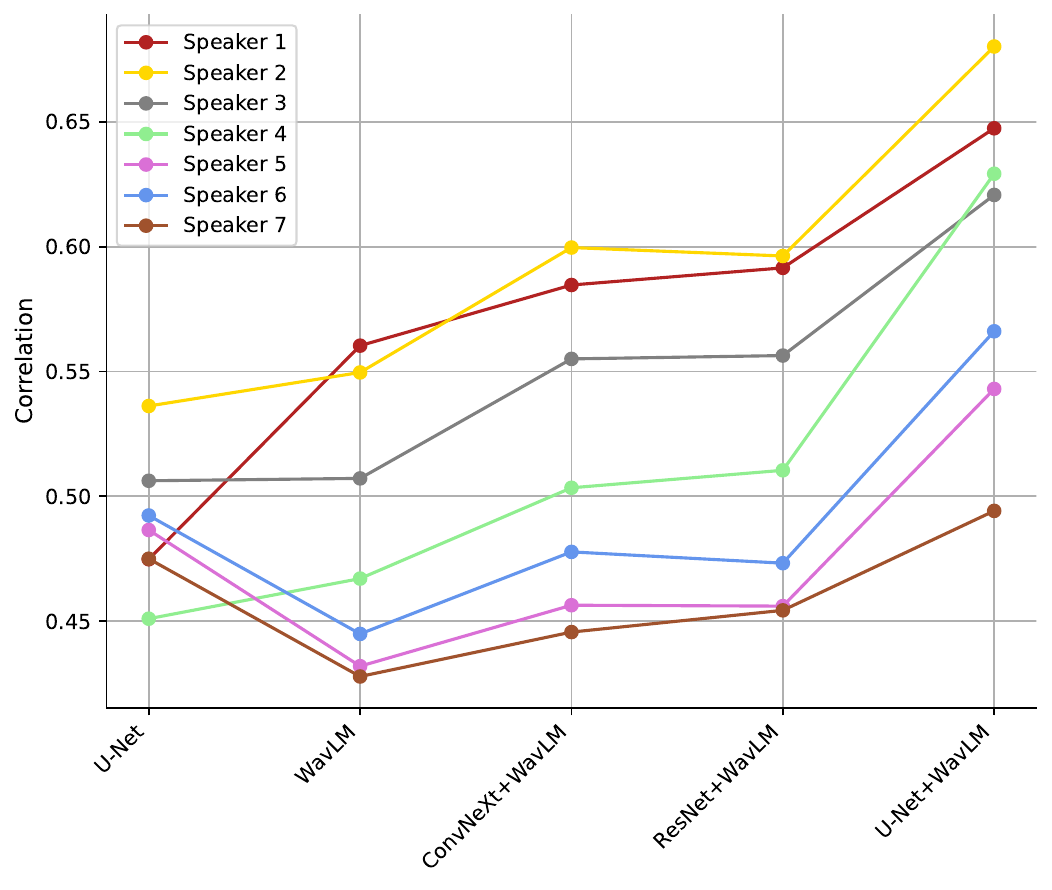}}
    \caption{L1 losses [↓] (\textit{left}) and Pearson
    Correlation Coefficients (PCCs) [↑] (\textit{right}) comparing 
    MRI trajectories of unseen examples from seen
    speakers of a given model with the USC-TIMIT ground truth. Varying
    through a subset of six representative models.}
    \label{fig:seg_results}
\end{figure*}

When analyzing our various feature extraction methods, we
first evaluate performance within the context of seen speakers but
unseen examples. Figure \ref{fig:seg_results} highlights quantitative results in L1 losses and Pearson
Correlation Coefficients (PCCs) when evaluating models on unseen examples from
seen speakers. We observe that multimodal models perform
consistently better than the purely video-based U-Net. In fact, the best model
in terms of both metrics includes the outputs of the U-Net as one of the input
modalities alongside WavLM vectors. These results suggest the inclusion of speech
within segmentation provides additional speaker-specific information related to the
anatomy of the vocal tract. Since the shape of different parts of the vocal tract can
greatly vary from speaker to speaker, this inclusion is crucial for better in-domain
modeling of speech production.
With only a single modality, the pixel value-based U-Net generalizes better
to unseen speakers than the WavLM-based speech inversion model since contour pixel values capture speaker-specific anatomy better than speech waveforms alone.

Similarly, we evaluate our segmentation methods on downstream speech tasks using speech synthesis 
within seen and unseen speaker contexts. Using the state-of-the-art MRI synthesis model \cite{synthesis2024} pretrained on
the newly-labeled 75-speaker dataset, we finetune on the projected MRI trajectories of a USC-TIMIT speaker provided by the different feature extraction models (i.e. baseline, U-Net, and multimodal). To evaluate the intelligibility of synthesized speech, we compute the word error
rate (WER) on test unseen examples using Whisper \cite{whisper}, a
state-of-the-art automatic speech recognition (ASR) model. For seen speakers, speech synthesized using the multimodal U-Net + WavLM based segmentations is more intelligible than speech synthesized from either
the ground truth baseline or the U-Net outputs, suggesting that the addition
of the speech modality helps preserves more speech-related information within the
predicted MRI point trajectories compared to a purely image-based approach.  Table \ref{tab:seen_speaker} 
summarizes these results. The results in Table~\ref{tab:unseen_speaker} highlight that
the U-Net + WavLM based model has the lowest WER when testing on an
unseen USC-TIMIT speaker, documenting that the segmentations from the multimodal model on unseen speakers still capture representative articulatory kinematics for naturalistic speech. Pretraining the synthesis model on the 75 speakers also results in much better unseen speaker 
generalization, demonstrating that the new labels for the Speech MRI Open Dataset are beneficial for 
future work in articulatory speech.

% \begin{table}[]
% \centering
% \caption{USC-TIMIT speaker finetuning for seen and 
% unseen speakers: 
% Mean WER for speech synthesis pretrained on 75-speaker 
% dataset. 
% (S) denotes synthesis model pretrained using single 
% MRI speaker. 
% All other models are pretrained with 75-speaker MRI.}
% \begin{tabular}{ccc}
% \toprule

% {} & \multicolumn{2}{c}{\textbf{Mean WER [↓]}}\\
% \textbf{Model} & \textbf{Seen Speaker} & 
% \textbf{Unseen Speaker}\\ \midrule
% U-Net + WavLM  & \textbf{0.31} $\pm$ 0.36 & 
% \textbf{0.33} $\pm$ 0.26 \\ \midrule
% U-Net          & 0.36 $\pm$ 0.33 & 0.35 $\pm$ 0.33 \\ 
% \midrule
% Ground Truth   & 0.34 $\pm$ 0.35 & 0.50 $\pm$ 0.27\\ 
% \midrule
%  U-Net + WavLM (S)&0.35 $\pm$ 0.33 & 0.50 $\pm$ 0.39\\ 
%  \bottomrule
% \end{tabular}
% \label{tab:synth}
% \end{table}

\begin{table}
\centering
\caption{Speech synthesis ASR WER finetuning on segmentations from a seen speaker during segmentation model training, but unseen utterances.
(S) denotes synthesis model pretrained using single 
MRI speaker. 
All other models are pretrained with 75-speaker MRI.}
\begin{tabular}{ll}
\toprule
            \textbf{Model} & \textbf{WER [\%]} \\
\midrule
    U-Net + WavLM & \textbf{31.3\% (16.4\%-49.3\%)} \\
            U-Net & 36.4\% (20.9\%-55.1\%) \\
     Ground Truth & 34.7\% (18.6\%-53.2\%) \\
U-Net + WavLM (S) & 34.9\% (20.3\%-52.8\%) \\
\bottomrule
\end{tabular}
\label{tab:seen_speaker}
\end{table}

\begin{table}
\centering
\caption{Speech synthesis ASR WER finetuning on segmentations from an unseen speaker during segmentation model training.
(S) denotes synthesis model pretrained using single 
MRI speaker. 
All other models are pretrained with 75-speaker MRI.}
\begin{tabular}{ll}
\toprule
            \textbf{Model} & \textbf{WER [\%]} \\
\midrule
    U-Net + WavLM & \textbf{33.3\% (20.2\%-49.8\%)} \\
            U-Net & 35.2\% (17.2\%-56.8\%) \\
     Ground Truth & 49.7\% (34.8\%-66.6\%) \\
U-Net + WavLM (S) & 50.1\% (28.0\%-72.8\%) \\
\bottomrule
\end{tabular}
\label{tab:unseen_speaker}
\end{table}

% \begin{table}
% \centering
% \caption{Speech synthesis ASR WER finetuning on segmentations from a seen speaker but unseen utterance during segmentation model training.
% (S) denotes synthesis model pretrained using single 
% MRI speaker. 
% All other models are pretrained with 75-speaker MRI.}
% \begin{tabular}{ll}
% \toprule
%             \textbf{Model} & \textbf{WER} \\
% \midrule
%     U-Net + WavLM & \textbf{0.313 (0.164-0.493)} \\
%             U-Net & 0.364 (0.209-0.551) \\
%      Ground Truth & 0.347 (0.186-0.532) \\
% U-Net + WavLM (S) & 0.349 (0.203-0.528) \\
% \bottomrule
% \end{tabular}

% \begin{tabular}{ll}
% \toprule
%             \textbf{Model} & \textbf{WER} \\
% \midrule
%     U-Net + WavLM & \textbf{0.333 (0.202-0.498)} \\
%             U-Net & 0.352 (0.172-0.568) \\
%      Ground Truth & 0.497 (0.348-0.666) \\
% U-Net + WavLM (S) & 0.501 (0.280-0.728) \\
% \bottomrule
% \end{tabular}
% \label{tab:unseen_speaker}
% \end{table}

\subsection{Qualitative Evaluation}

Despite relying on the output of the baseline segmentation algorithm as the training targets, 
our segmentation methods performed better than the baseline algorithm when 
evaluated on downstream speech synthesis. We hypothesize that this is because the baseline 
segmentations have high amounts of jitter and inconsistencies across frames, and are sometimes even physiologically implausible. In comparison, the estimates of the deep learning approaches do not have the same level of frame-dependent noise, possibly explaining why they are better suited for 
building downstream methods. To validate this hypothesis with a subjective evaluation, we ran a 
one-tailed perceptual test for statistical significance where 
participants looked at two video animations of vocal tract movements in 
side-by-side panels (one with the baseline labels, and the other with outputs 
of our segmentation method). The participants then selected which rendering 
is a more accurate representation of the associated audio. Each participant repeated this process for five test examples. Our 
results reveal the participants (\textit{n=}21) 
prefer the outputs of our algorithm over the baseline 
segmentations (p $<$ 0.001). For visualization of these results, we invite you to
watch our demo video at \url{rishiraij.github.io/multimodal-mri-avatar}.

\section{Conclusion}
\label{sec:conclusion}

In this work, we looked at developing a generalizable articulatory segmentation algorithm from RT-MRI videos of the vocal tract. We used the limited existing articulatory labeling to train vision-based and multimodal models which efficiently and accurately extract physiological features from MRI videos of unseen speakers. Through speech synthesis, we demonstrate that our approach results in higher quality segmentations for downstream speech tasks than existing baselines, while also being more accurate representations of speech audio. While MRI-based articulatory modeling is less studied than other approaches such as EMA, we hope that our released labeling of 75 speakers will allow future work in speech modeling and linguistics to take advantage of the more-complete physiological representation that RT-MRI provides.

% \section{Acknowledgements}
% This research is supported by the following grants to PI Anumanchipalli — NSF award 2106928, Google Research Scholar Award, Rose Hills Foundation and Noyce Foundation.

% \ifinterspeechfinal
%      The Interspeech 2024 organisers
% \else
%      The authors
% \fi
% would like to thank ISCA and the organising committees of past Interspeech conferences for their help and for kindly providing the previous version of this template.

\bibliographystyle{IEEEtran}
\bibliography{mybib}

\end{document}